%% file: main.tex
\documentclass[letterpaper, 10 pt, journal, twoside]{IEEEtran}
\usepackage{graphics} 
\usepackage{epsfig} 

\usepackage{times} 
\usepackage{amsmath} 
\usepackage{amssymb} 
\usepackage{bm}
\usepackage{amsfonts}
\DeclareMathAlphabet{\mathpzc}{OT1}{pzc}{m}{it}

\usepackage{hyperref}
\usepackage[capitalise,nameinlink]{cleveref}
\usepackage{color}
\usepackage{tabularx}
\usepackage{siunitx}

\usepackage{float}
\usepackage{caption}
\usepackage{subcaption}
\usepackage{xspace}
\newcommand{\eg}{\hbox{\emph{e.g.}}\xspace}

\usepackage{svg}

\ifCLASSINFOpdf
\else
\fi
\begin{document}
%
\title{Decentralized Ability-Aware Adaptive Control for Multi-robot Collaborative Manipulation}
%
\author{Lei Yan, Theodoros Stouraitis, and Sethu Vijayakumar%
\thanks{Manuscript received: October, 15, 2020; Revised December, 31, 2020; Accepted January, 25, 2021. This letter was recommended for publication by Associate Editor M. Ani Hsieh and Editor A. M. Okamura upon evaluation of the reviewers' comments. This work is supported by 
EPSRC UK RAI Hub in Future AI and Robotics for Space (FAIR-SPACE: EP/R026092/1), Shenzhen Institute of Artificial Intelligence and Robotics for Society and the Honda Research Institute Europe. \textit{(Corresponding author: Sethu Vijayakumar.)}} 
\thanks{Lei Yan, Theodoros Stouraitis and Sethu Vijayakumar are with the Institute of Perception, Action and Behaviour, School of Informatics, University of Edinburgh, Edinburgh, EH8 9AB, U.K. (e-mail: lei.yan@ed.ac.uk, theodoros.stouraitis@ed.ac.uk, sethu.vijayakumar@ed.ac.uk).}
\thanks{Sethu Vijayakumar is a visiting researcher at the Shenzhen Institute for Artificial Intelligence and Robotics for Society (AIRS).}%
\thanks{Digital Object Identifier: see top of this page.}
}
\markboth{IEEE Robotics and Automation Letters. Preprint Version. Accepted January, 2021}
{Yan \MakeLowercase{\textit{et al.}}: Decentralized ability-aware adaptive control} 

\maketitle
\begin{abstract}
Multi-robot teams can achieve more dexterous, complex and heavier payload tasks than a single robot, yet effective collaboration is required. Multi-robot collaboration is extremely challenging due to the different kinematic and dynamics capabilities of the robots, the limited communication between them, and the uncertainty of the system parameters. In this paper, a Decentralized Ability-Aware Adaptive Control ($\textit{DA}^3\textit{C}$) is proposed to address these challenges based on two key features. Firstly, the common manipulation task is represented by the proposed nominal task ellipsoid, which is used to maximize each robot's force capability online via optimizing its configuration. Secondly, a decentralized adaptive controller is designed to be Lyapunov stable in spite of heterogeneous actuation constraints of the robots and uncertain physical parameters of the object and environment. In the proposed framework, decentralized coordination and load distribution between the robots is achieved without communication, while only the control deficiency is broadcast if any of the robots reaches its force limits. In this case, the object's reference trajectory is modified in a decentralized manner to guarantee stable interaction. Finally, we perform several numerical and physical simulations to analyse and verify the proposed method with heterogeneous multi-robot teams in collaborative manipulation tasks.
\end{abstract}


\section{Introduction}\label{sec:intro}

\input{sections/intro.tex}

\section{Preliminaries}\label{sec:rel_work}
\input{sections/preliminary.tex}

\section{Task-oriented Null-space Manipulability Optimization}\label{sec:task_oriented_optimization}

\input{sections/task_oriented_optimization.tex}
\section{Decentralized Ability-Aware \\ Adaptive Control} 
\label{sec:load_distribution}
\input{sections/A3C.tex}

\section{Results}\label{sec:results}

\input{sections/results.tex}
\section{Conclusion}\label{sec:conclusion}

\input{sections/conclusion.tex}




%

\appendices
\input{sections/appendix.tex}




\ifCLASSOPTIONcaptionsoff
  \newpage
\fi



%





\bibliographystyle{IEEEtran}
\bibliography{reference}  

\addtolength{\textheight}{-12cm}   


\end{document}

%% file: sections/intro.tex
\begin{figure}[t]
    \begin{center}
        \fontsize{8pt}{11pt}
        \def\svgwidth{0.60\columnwidth}
        \input{figures/multiRobotConcept_index.tex}
    	\vspace{-10pt}
    \end{center}
    \caption{Pictorial description of the multi-robot collaborative manipulation setup, where the ability of each robot is illustrated as a force polytope.}
    \label{fig:scenario}
   	\vspace{-14pt}
\end{figure}
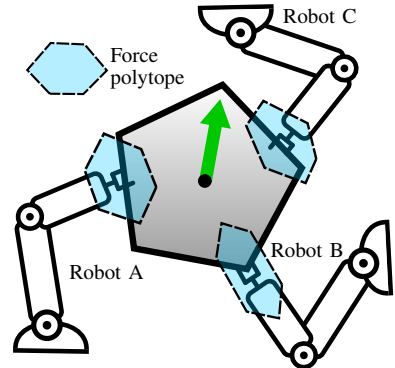

Collaboration with other agents can often be beneficial.
For example, a multi-robot team like the one shown in~Fig.~\ref{fig:scenario} is more dexterous and robust in heavy and large object manipulation tasks~\cite{culbertson2018decentralized} than a single robot.
Also, in human-robot collaboration scenarios~\cite{Stou2020Online}, the human's input can improve the intelligence and adaptability of the team. Yet, collaboration is not trivial, due to the effects of one agent's actions on the planning, control and decision of others.

Here, we investigate multi-robot collaborative manipulation tasks, where a decentralized robot team needs to achieve a common objective, while each robot has different motion and force capabilities. 
To perform such collaborative tasks, each robot within the team should maximize its contribution to the task, 
appropriately distribute the load among other robots,
and adapt its behaviour according to the capabilities of the other robots in the team.
Traditional centralized control methods have been used for multi-robot collaboration, but assume access to an accurate model of the robot team and full observability of the state of the other robots and the object.
However, when considering the characteristics of real-world multi-robot teams~\cite{emam2020adaptive}, the most pressing problems are: (i) heterogeneity of the robots' capabilities, (ii) uncertainty of the system's physical parameters and (iii) lack of high bandwidth communication between the robots. 

Therefore, we propose a method to maximize the force capability of each robot while designing a decentralized adaptive controller. Using this framework, we achieve the shared manipulation task under modelling uncertainties, input constraints and band-limited communication. 



\textit{Motion and force capabilities:}
A manipulability metric~\cite{yoshikawa1985manipulability_J} was first proposed as a measure of the capability of robotic mechanisms, and has been broadly used for redundancy control~\cite{yoshikawa1985manipulability_C}.
Utilizing the task-oriented manipulability measure, the optimal joint configuration of a redundant manipulator can be determined~\cite{lee1989dual}. An efficient closed-form calculation of the task space manipulability was presented for a 7-DOF 
manipulator~\cite{huber2019efficient}. 
For a multiple-arm 
system, the task-space force and velocity manipulability ellipsoids were given in~\cite{chiacchio1991global}. 
Based on a study of the 
dynamic manipulability of robots, two physically meaningful choices for weighting matrix were provided~\cite{azad2019effects}. To simplify the calculation of the dynamic manipulability, the weighted manipulability ellipsoid can be used to approximate the manipulability polytope~\cite{chiacchio1997force}. In this paper, the weighted force manipulability ellipsoid (WFME) will be adopted to optimize the force polytope of the manipulator.

\begin{figure}[!t]
    \begin{center}
        \vspace{1.5mm}
        \fontsize{8.5pt}{10pt}
        \def\svgwidth{0.955\columnwidth}
        \input{figures/multiRobotPipeline_index.tex}
    	\vspace{-5pt}
    \end{center}
    \caption{Flowchart of decentralized ability-aware adaptive control.}
    \label{fig:A3C_flowchart}
   	\vspace{-15pt}
\end{figure}
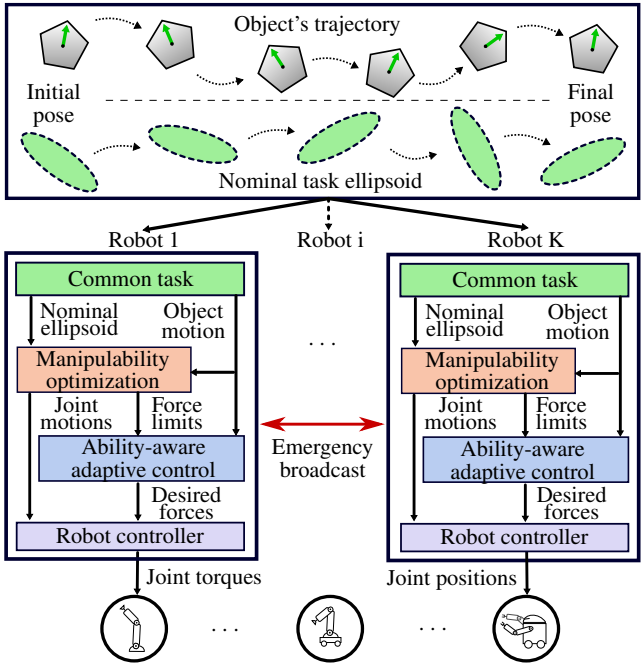

\textit{Redundancy exploitation:}
For dexterous manipulation~\cite{yan2016coordinated, yan2018dual}, redundant manipulators were 
adopted to enlarge workspace, avoid singularities and collisions with the environment. Further, manipulator's redundancy can also be explored to maximize manipulability~\cite{jin2017manipulability}. 
The stiffness feasibility regions of redundant manipulators 
and a global task-oriented stiffness optimization 
were used to select robot poses~\cite{ajoudani2017choosing} and configurations~\cite{busson2017task} for force and stiffness control, respectively. 
This highlights the important role of the robot's configuration in interaction tasks. The configuration of a redundant robot was also 
optimized while manipulating an object in space, to minimize disturbances on its base~\cite{yan2020multi}. 
Geometry-aware methods were used 
to provide a manipulability-based redundancy resolution~\cite{rozo2017learning, jaquier2018geometry}, which enabled tracking of manipulability ellipsoids. Yet, the desired ellipsoid is either pre-defined by a human expert or pre-recorded from demonstrations.
Here, we define a nominal task ellipsoid which is generated from the manipulation task automatically and present a task-oriented manipulability optimization to match the WFME of the robot with the nominal task ellipsoid.

\textit{Decentralized adaptive control:}
A decentralized model reference adaptive controller was proposed to deal with uncertain physical parameters of the collaborative multi-robot system~\cite{liu1998decentralized}.  For manipulation tasks with inaccurate kinematic model, the adaptive controller was used to handle the closed kinematic chain constraint and achieve accurate motion tracking with minimum-norm actuation force~\cite{aghili2012adaptive}. Recently, Deep Neural Networks were adopted to model system uncertainties in model reference adaptive control~\cite{Joshi2019deep}.
In studies of multiple mobile manipulators, distributed coordination control and synchronous cooperation control were presented to deal with time delays and switching topologies~\cite{dai2016distributed}. A distributed cooperation scheme was adopted for networked mobile manipulators, which exploits the formation-based task allocation and task-oriented strategy~\cite{ren2020fully}. A distributed
impedance controller has also been used for collaborative manipulation with event-triggered  communication~\cite{dohmann2020distributed}. Recently, a decentralized adaptive controller~\cite{culbertson2018decentralized} for multiple collaborative mobile robots was introduced. This controller can track the reference velocity trajectory without a priori knowledge of the agent's position and payload properties. Yet, all adaptive controllers described above did not consider force and torque constraints (manipulability) of the robots.

Considering a multi-robot collaborative manipulation task, a decentralized ability-aware adaptive control ($\textit{DA}^3\textit{C}$) framework (shown in Fig.~\ref{fig:A3C_flowchart}) is proposed. Our method can handle both uncertain system parameters and input constraints without full communication between the robots. 
In the investigated multi-robot collaborative manipulation setup each robot has access to the desired manipulation task, which is described as a nominal task ellipsoid. Each robot tracks the nominal task ellipsoid using the task-oriented manipulability optimization method, while the $\textit{DA}^3\textit{C}$ enables multi-robot coordination with respect to the common manipulation task.
The main contributions of this paper are summarised as follows: 
\begin{enumerate}
	\item A nominal task ellipsoid is defined based on the common manipulation task, and it is used to optimize the force capability of each manipulator.
	
	\item A decentralized adaptive controller under input constraints is designed and proven to be Lyapunov stable.
	
	
\item Different heterogeneous multi-robot systems with input and communication constraints realize
collaborative manipulation tasks using the proposed decentralized ability-aware adaptive control that guarantees stability and convergence.
\end{enumerate}

The remainder of this paper is organized as follows. The preliminary work is presented in Section~\ref{sec:rel_work}. In Section~\ref{sec:task_oriented_optimization}, we define the nominal task ellipsoid and present the task-oriented manipulability optimization. The decentralized ability-aware adaptive control is described in Section~\ref{sec:load_distribution}. In depth analysis of the proposed method is carried out in Section~\ref{sec:results}, where several numerical and physical simulations of collaborative manipulation tasks are performed. 
The conclusion and future work are discussed in Section~\ref{sec:conclusion}.


%% file: figures/multiRobotConcept_index.tex
\begingroup%
  \makeatletter%
  \providecommand\color[2][]{%
    \errmessage{(Inkscape) Color is used for the text in Inkscape, but the package 'color.sty' is not loaded}%
    \renewcommand\color[2][]{}%
  }%
  \providecommand\transparent[1]{%
    \errmessage{(Inkscape) Transparency is used (non-zero) for the text in Inkscape, but the package 'transparent.sty' is not loaded}%
    \renewcommand\transparent[1]{}%
  }%
  \providecommand\rotatebox[2]{#2}%
  \newcommand*\fsize{\dimexpr\f@size pt\relax}%
  \newcommand*\lineheight[1]{\fontsize{\fsize}{#1\fsize}\selectfont}%
  \ifx\svgwidth\undefined%
    \setlength{\unitlength}{320.31496063bp}%
    \ifx\svgscale\undefined%
      \relax%
    \else%
      \setlength{\unitlength}{\unitlength * \real{\svgscale}}%
    \fi%
  \else%
    \setlength{\unitlength}{\svgwidth}%
  \fi%
  \global\let\svgwidth\undefined%
  \global\let\svgscale\undefined%
  \makeatother%
  \begin{picture}(1,0.95575221)%
    \lineheight{1}%
    \setlength\tabcolsep{0pt}%
    \put(0,0){\includegraphics[width=\unitlength,page=1]{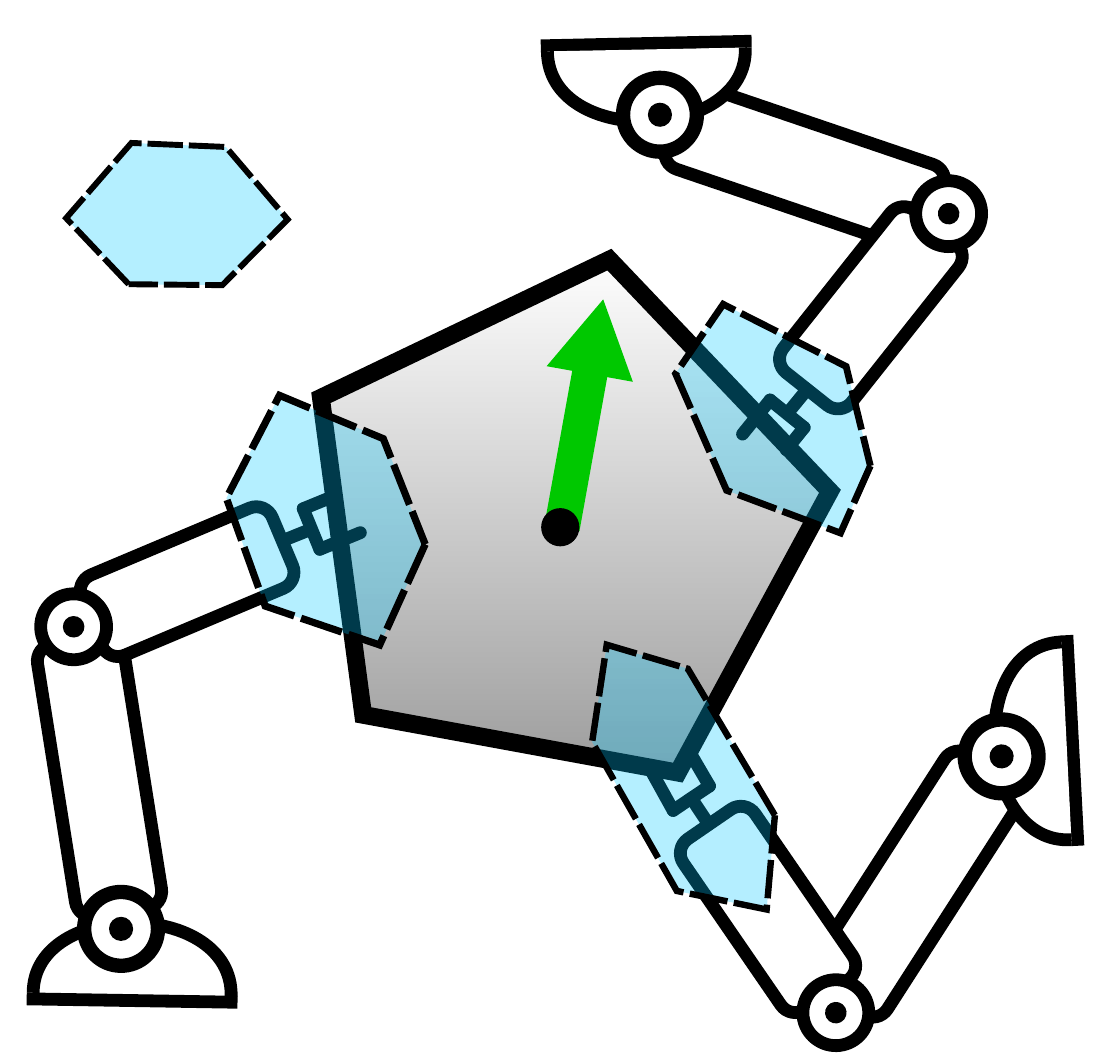}}%
    \put(0.2678201,0.78017499){\color[rgb]{0,0,0}\makebox(0,0)[lt]{\lineheight{1.05}\smash{\begin{tabular}[t]{l}Force\\polytope\end{tabular}}}}%
    \put(0.16506436,0.23550261){\color[rgb]{0,0,0}\makebox(0,0)[lt]{\lineheight{1.25}\smash{\begin{tabular}[t]{l}Robot A\end{tabular}}}}%
    \put(0.66847652,0.29090268){\color[rgb]{0,0,0}\makebox(0,0)[lt]{\lineheight{1.25}\smash{\begin{tabular}[t]{l}Robot B\end{tabular}}}}%
    \put(0.69989154,0.87605537){\color[rgb]{0,0,0}\makebox(0,0)[lt]{\lineheight{1.25}\smash{\begin{tabular}[t]{l}Robot C\end{tabular}}}}%
  \end{picture}%
\endgroup%

%% file: figures/multiRobotPipeline_index.tex
\begingroup%
  \makeatletter%
  \providecommand\color[2][]{%
    \errmessage{(Inkscape) Color is used for the text in Inkscape, but the package 'color.sty' is not loaded}%
    \renewcommand\color[2][]{}%
  }%
  \providecommand\transparent[1]{%
    \errmessage{(Inkscape) Transparency is used (non-zero) for the text in Inkscape, but the package 'transparent.sty' is not loaded}%
    \renewcommand\transparent[1]{}%
  }%
  \providecommand\rotatebox[2]{#2}%
  \newcommand*\fsize{\dimexpr\f@size pt\relax}%
  \newcommand*\lineheight[1]{\fontsize{\fsize}{#1\fsize}\selectfont}%
  \ifx\svgwidth\undefined%
    \setlength{\unitlength}{314.81008186bp}%
    \ifx\svgscale\undefined%
      \relax%
    \else%
      \setlength{\unitlength}{\unitlength * \real{\svgscale}}%
    \fi%
  \else%
    \setlength{\unitlength}{\svgwidth}%
  \fi%
  \global\let\svgwidth\undefined%
  \global\let\svgscale\undefined%
  \makeatother%
  \begin{picture}(1,1.03752479)%
    \lineheight{1}%
    \setlength\tabcolsep{0pt}%
    \put(0,0){\includegraphics[width=\unitlength,page=1]{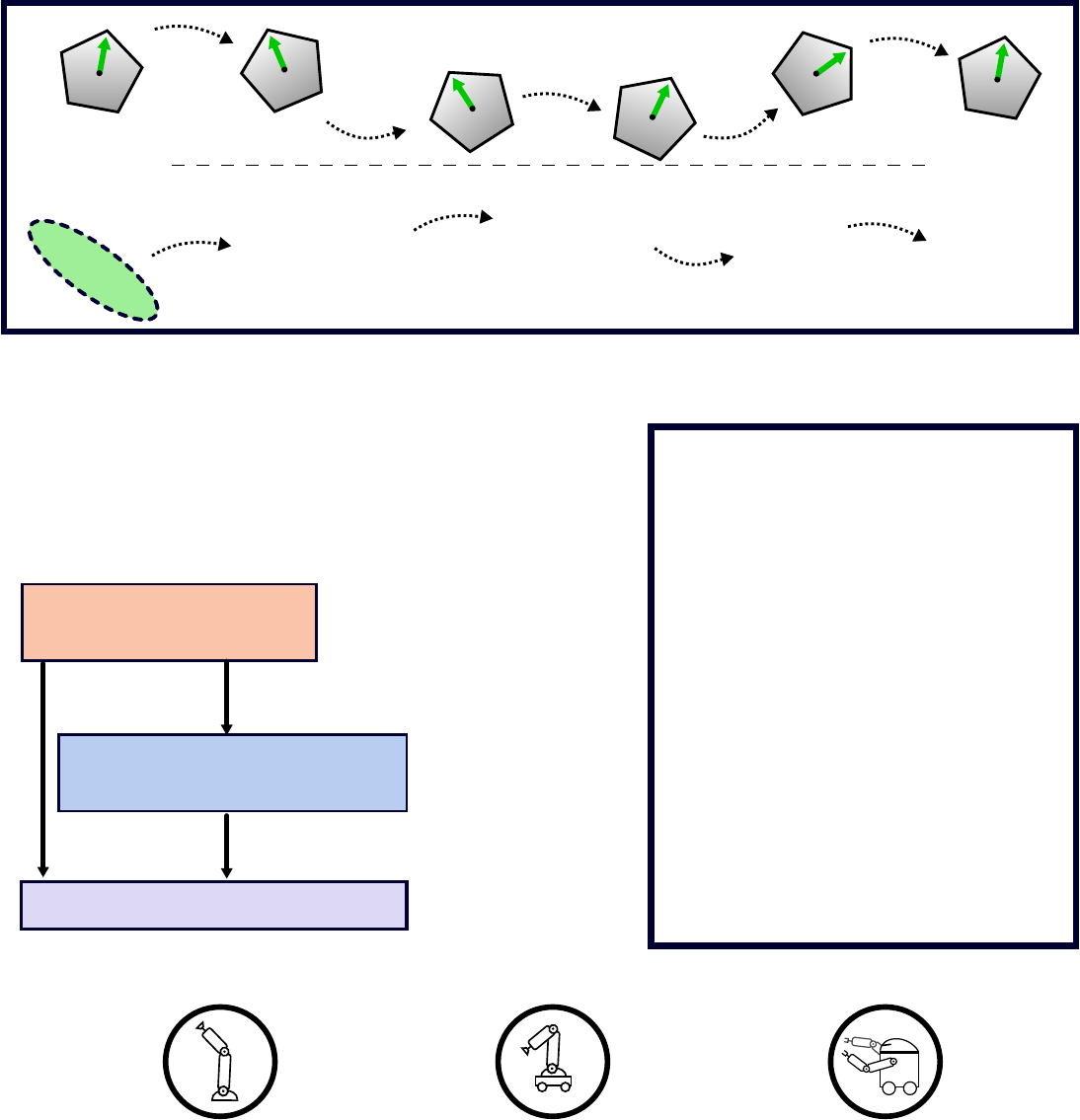}}%
    \put(0.32501492,0.05030095){\color[rgb]{0,0,0}\makebox(0,0)[lt]{\lineheight{1.25}\smash{\begin{tabular}[t]{l}. . .\end{tabular}}}}%
    \put(0.65163781,0.05062712){\color[rgb]{0,0,0}\makebox(0,0)[lt]{\lineheight{1.25}\smash{\begin{tabular}[t]{l}. . .\end{tabular}}}}%
    \put(0,0){\includegraphics[width=\unitlength,page=2]{figures/multiRobotPipeline.pdf}}%
    \put(0.47795508,0.49902294){\color[rgb]{0,0,0}\makebox(0,0)[lt]{\lineheight{1.25}\smash{\begin{tabular}[t]{c}. . .\end{tabular}}}}%
    \put(0,0){\includegraphics[width=\unitlength,page=3]{figures/multiRobotPipeline.pdf}}%
    \put(0.08138412,0.18792077){\color[rgb]{0,0,0}\makebox(0,0)[lt]{\smash{\begin{tabular}[t]{l}Robot controller\end{tabular}}}}%
    \put(0.4206905,0.32770512){\color[rgb]{0,0,0}\makebox(0,0)[lt]{\smash{\begin{tabular}[t]{c}Emergency\\broadcast\end{tabular}}}}%
    \put(0.10058174,0.59195704){\color[rgb]{0,0,0}\makebox(0,0)[lt]{\smash{\begin{tabular}[t]{l}Common task\end{tabular}}}}%
    \put(0.49217875,0.99362805){\color[rgb]{0,0,0}\makebox(0,0)[t]{\smash{\begin{tabular}[t]{c}Object's trajectory\end{tabular}}}}%
    \put(0.03554796,0.88572672){\color[rgb]{0,0,0}\makebox(0,0)[lt]{\smash{\begin{tabular}[t]{c}Initial \\pose\end{tabular}}}}%
    \put(0.88572955,0.88572672){\color[rgb]{0,0,0}\makebox(0,0)[lt]{\smash{\begin{tabular}[t]{c}Final\\pose\end{tabular}}}}%
    \put(0.4973371,0.7465116){\color[rgb]{0,0,0}\makebox(0,0)[t]{\smash{\begin{tabular}[t]{c}Nominal task ellipsoid\end{tabular}}}}%
    \put(0.05012039,0.4671767){\color[rgb]{0,0,0}\makebox(0,0)[lt]{\smash{\begin{tabular}[t]{c}Manipulability\\optimization\end{tabular}}}}%
    \put(0.10530955,0.32732874){\color[rgb]{0,0,0}\makebox(0,0)[lt]{\smash{\begin{tabular}[t]{c}Ability-aware\\adaptive control\end{tabular}}}}%
    \put(0.05648936,0.39429383){\color[rgb]{0,0,0}\makebox(0,0)[lt]{\lineheight{0.25}\smash{\begin{tabular}[t]{c}Joint\\motions\end{tabular}}}}%
    \put(0.23152092,0.39429383){\color[rgb]{0,0,0}\makebox(0,0)[lt]{\lineheight{0.25}\smash{\begin{tabular}[t]{c}Force\\limits\end{tabular}}}}%
    \put(0.23152092,0.25242499){\color[rgb]{0,0,0}\makebox(0,0)[lt]{\lineheight{0.25}\smash{\begin{tabular}[t]{c}Desired\\forces\\\end{tabular}}}}%
    \put(0.75973847,0.65353106){\color[rgb]{0,0,0}\makebox(0,0)[lt]{\smash{\begin{tabular}[t]{l}Robot K\end{tabular}}}}%
    \put(0.16363307,0.65353092){\color[rgb]{0,0,0}\makebox(0,0)[lt]{\smash{\begin{tabular}[t]{l}Robot 1\end{tabular}}}}%
    \put(0.2236659,0.12294679){\color[rgb]{0,0,0}\makebox(0,0)[lt]{\smash{\begin{tabular}[t]{l}Joint torques\end{tabular}}}}%
    \put(0.60105268,0.12265578){\color[rgb]{0,0,0}\makebox(0,0)[lt]{\smash{\begin{tabular}[t]{l}Joint positions\end{tabular}}}}%
    \put(0.4603216,0.65353092){\color[rgb]{0,0,0}\makebox(0,0)[lt]{\smash{\begin{tabular}[t]{l}Robot i\end{tabular}}}}%
    \put(0.24725953,0.54271255){\color[rgb]{0,0,0}\makebox(0,0)[lt]{\lineheight{0}\smash{\begin{tabular}[t]{c}Object\\motion\end{tabular}}}}%
    \put(0.05648936,0.54271255){\color[rgb]{0,0,0}\makebox(0,0)[lt]{\lineheight{0.25}\smash{\begin{tabular}[t]{c}Nominal\\ellipsoid\end{tabular}}}}%
    \put(0.70464532,0.59009589){\color[rgb]{0,0,0}\makebox(0,0)[lt]{\smash{\begin{tabular}[t]{l}Common task\end{tabular}}}}%
    \put(0.6621838,0.46531541){\color[rgb]{0,0,0}\makebox(0,0)[lt]{\smash{\begin{tabular}[t]{c}Manipulability\\optimization\end{tabular}}}}%
    \put(0.70137306,0.32446732){\color[rgb]{0,0,0}\makebox(0,0)[lt]{\smash{\begin{tabular}[t]{c}Ability-aware\\adaptive control\end{tabular}}}}%
    \put(0.68544763,0.18605935){\color[rgb]{0,0,0}\makebox(0,0)[lt]{\smash{\begin{tabular}[t]{l}Robot controller\end{tabular}}}}%
    \put(0.66055298,0.39443254){\color[rgb]{0,0,0}\makebox(0,0)[lt]{\lineheight{0.25}\smash{\begin{tabular}[t]{c}Joint\\motions\end{tabular}}}}%
    \put(0.83558437,0.39424404){\color[rgb]{0,0,0}\makebox(0,0)[lt]{\lineheight{0.25}\smash{\begin{tabular}[t]{c}Force\\limits\end{tabular}}}}%
    \put(0.83685164,0.25242499){\color[rgb]{0,0,0}\makebox(0,0)[lt]{\lineheight{0.25}\smash{\begin{tabular}[t]{c}Desired\\forces\end{tabular}}}}%
    \put(0.85132308,0.54013544){\color[rgb]{0,0,0}\makebox(0,0)[lt]{\lineheight{0.25}\smash{\begin{tabular}[t]{c}Object\\motion\end{tabular}}}}%
    \put(0.66220839,0.5408514){\color[rgb]{0,0,0}\makebox(0,0)[lt]{\lineheight{0.25}\smash{\begin{tabular}[t]{c}Nominal\\ellipsoid\end{tabular}}}}%
  \end{picture}%
\endgroup%

%% file: sections/preliminary.tex


\subsection{Manipulability Ellipsoid and Force Polytope}
Generally speaking, the force manipulability ellipsoid can be used to approximately describe the force capability of the manipulator. In the simplest case, if we consider an arbitrary $n$-DOF manipulator robot $k$ with the same torque limit across all its joints, 
we can obtain the force manipulability ellipsoid~\cite{yoshikawa1985manipulability_J} using joint torque ${\bm \tau_k}$ such that $\|{\bm \tau_k}\|_2 \leqslant 1$. This ellipsoid is a subset of all realizable forces and is defined as
\begin{equation}\label{eqn:force_manipulability_ellipsoid}
{\bm F}_k^T \left( \bm J_k \bm J_k^T\right)  {\bm F_k} \leqslant 1,
\end{equation}
where ${\bm J_k}$ is the Jacobian matrix of robot $k$ and ${\bm F_k}$ is the force (and torque) at the end-effector of robot $k$.

However, typically manipulators have different torque limits for each joint, expressed as $|\tau_k^i| \leqslant \tau_{k,max}^i$ for $i = 1,...,n$, where $\tau_{k,max}^i$ is the maximum torque of $i$-th joint of robot $k$. Thus, the force polytope~\cite{chiacchio1997force} of the manipulator is described by 2$n$ bounding inequalities as
\begin{equation}\label{eqn:torque_limit}
-\bm {\tau}_{k,max} \leqslant \bm J_k^T  {\bm F_k} \leqslant \bm {\tau}_{k,max}.
\end{equation}

As shown on the left side of Fig.~\ref{fig:singlearm_illustration}, the force polytope corresponding to joint torque limits in  \eqref{eqn:torque_limit} can be approximated by the weighted force manipulability ellipsoid (WFME) as 

\begin{equation}
	\bm F_k^T \bm J_k \bm W_k^T \bm W_k \bm J_k^T \bm F_k \leqslant 1,
\end{equation}
where $\bm W_k = \text{diag}\left(\frac{1}{\tau_{k,max}^1},\cdots, \frac{1}{\tau_{k,max}^n}\right)$ is a 
weighting matrix used to formulate the WFME.

\subsection{Object's Dynamics}
For multi-robot collaborative manipulation scenarios, the equation of motion of the object can be written as 
\begin{equation}\label{eqn:target_dynamic_equation}
{\left[ \begin{array}{cc} m_o \dot{\bm v}_o \\ \bm I_o \dot{\bm \omega}_o+ \bm \omega_o \times({\bm I_o}{\bm \omega_o}) \end{array} \right ]} = {\bm F_t}+ {\bm F_f}+{\bm G_o}, 
\end{equation}
where ${{m}_o}$ and $\bm I_o$ are the mass and inertia of the object, ${{\bm v}_o}$ and $\bm \omega_o$ are respectively the linear velocity and angular velocity of the object, ${\bm G_o} = [m_o \bm g^T,\ {\bm 0}]^T$ is the gravity, ${\bm F_f}$ is the linear and rotational friction force which is modelled as
$\bm F_f = [\bm f_l^T, \bm f_r^T]^T=[-\mu_l \bm v_t^T, -\mu_r \bm \omega_t^T]^T$, $\mu_l$ and $\mu_r$ are the linear and rotational friction coefficients, respectively.
${\bm F_t}$ is the external force exerted on the object (in this paper it is task-related).
All the variables are expressed in the world coordinate system.


%% file: sections/task_oriented_optimization.tex
To utilize each robot's maximum capability, its force polytope needs to be optimized with respect to the manipulation task. As WFME is a conservative approximation of the force polytope as shown in~\cref{fig:singlearm_illustration}, we optimize WFME instead to maximize each robot's force capability. To do this, we first define the nominal task ellipsoid to encode the task's force characteristics. 
Second, we optimize the robot's null-space motion to match the WFME with the nominal task ellipsoid. 

\subsection{Nominal Task Ellipsoid}\label{sec:nominal_task_ellipsoid}

The nominal task ellipsoid is defined as an ellipsoid of revolution, also called a prolate spheroid, whose two principle axes have the same length, while the third principle axis is the longest.
The nominal task ellipsoid can be generated by the transformation of unit sphere as
\begin{equation} \label{eqn:ellipsoid_representation}
\left\lbrace \bm c_e = \bm C \bm c_s \enspace | \enspace \|\bm c_s\| \leqslant 1\right\rbrace,  
\end{equation}
where $\bm c_e$ is the Cartesian coordinates of the nominal task ellipsoid, $\bm c_s$ is the Cartesian coordinates of the unit sphere. $\bm C$ is the transformation matrix and can be calculated as $\bm C = \bm R_r\bm R_s$
with $\bm R_s$ being the scaling matrix and $\bm R_r$ being the rotation matrix. The scaling matrix $\bm R_s$ is defined as $\bm R_s = \operatorname{diag} \left( l_x, \ l_y, \ l_z  \right)$,
where $l_x$, $l_y$, $l_z$, are the lengths of the principle axes. In this paper, $l_x=1$ corresponds to the longest axis that is aligned with 
the desired force. $l_y=l_z= c_f l_x$ correspond to the other two axes and $c_f \in \left(0, 1\right]$ can be set according to the task requirements, \eg $c_f=1$ (isotropic ellipsoid in Fig.~3) for manipulation tasks that require force to be equally distributed along different directions, while $c_f<1$ (other cases in Fig.~3) for manipulation tasks that require force along a specific direction.

Given the desired force $\bm F_t$ on the object
and the unit vector $\bm a_x$ of the longest axis, 
the common perpendicular vector $\bm u$ and the included angle $\phi$ between them, can be obtained by 
\begin{equation}
\bm u  = \dfrac{\bm F_t \times \bm a_x}{\| \bm F_t \times \bm a_x \|} \hspace{1.5mm} \text{and} \hspace{1mm}\ 
\phi  = acos \left( \dfrac {\bm F_t\cdot \bm a_x} {\| \bm F_t \| \| \bm a_x \|} \right),
\end{equation}
respectively. Therefore, the rotation matrix $\bm R_r$---which is used to align the longest axis with the desired force---
can be obtained according to the angle-axis representation as
\begin{equation}
\bm R_r = \bm E_3c_{\phi} + (1-c_{\phi})\bm u \bm u^T + \bm u^\times  s_{\phi},
\end{equation}
where $s_{\phi}=\sin(\phi)$, $c_{\phi}=\cos(\phi)$, $\bm E_3$ is the $3\times 3$ identity matrix and $\bm u^\times$ is the skew symmetric matrix of $\bm u$.  



\begin{figure}[!t]
    \begin{center}
        \vspace{-7pt}
        \def\svgwidth{0.85\columnwidth}
        \input{figures/force_polytope_illustration_index.tex}
    	\vspace{-8pt}
    \end{center}
	\caption{Force polytope and WFME (left). Nominal task ellipsoids corresponding to different manipulation tasks where $F_t$ represents the desired force along different directions (right).}\label{fig:singlearm_illustration}
	\vspace{-12pt}
\end{figure}
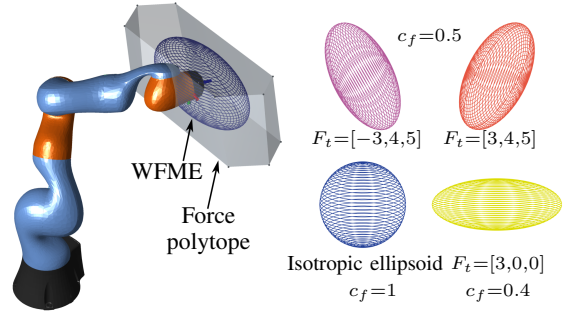

From \eqref{eqn:ellipsoid_representation}, we can obtain the following equation
\begin{equation}\label{eqn:nominal_task_ellipsoid}
{\bm c_e^T \bm M_t^{-1} \bm c_e \leqslant 1, \  \text{with }
\bm{M}_{t} = \bm C \bm C^T,}
\end{equation}
where the symmetric positive definite matrix $\bm{M}_{t}$ represents the nominal task ellipsoid. A few nominal task ellipsoids for different manipulation tasks are visualized in~\cref{fig:singlearm_illustration}. The shape of a nominal task ellipsoid is determined by the coefficient $c_f$, while its orientation is determined by the direction of the desired force $\bm F_t$.

\subsection{Null-Space Manipulability Optimization}\label{sec:single-arm optimization}
As multiple robots manipulate the object jointly, the end-effector of each robot can be assumed to be fixed on the object via the corresponding grasping point. Thus, the velocity of the end-effector of each robot ${\dot{\bm x}_k}$ can be derived from the velocity of the object ${\dot{\bm x}_o}$ as
\begin{equation}\label{eqn:pre vel constraint between target and arms}
{\dot{\bm x}_{k}}=\left[ \begin{array}{cc} 
\bm E_3 & -\bm r_k^\times \\ {\bm 0} & \bm E_3
\end{array} \right] {\dot{\bm x}_o }={\bm G_{pk}^T}{\dot{\bm x}_o},
\end{equation}
where $\bm G_{pk}$ is the grasp matrix of robot $k$, $\bm r_k$ is the position vector from the object's center of mass to the grasping point of robot $k$, and $\dot{\bm x}_o=\left[ \bm v_o^T, \ \bm \omega_o^T \right]^T$. By differentiating \eqref{eqn:pre vel constraint between target and arms} the acceleration constraint can be obtained as
\begin{equation}\label{eqn:pre acc constraint between target and arms}
{\ddot{\bm x}_{k}}={\bm G_{pk}^T}{\ddot{\bm x}_o }+{\dot{\bm G}_{pk}^T}{\dot{\bm x}_o }.
\end{equation}
At the same time, the pseudo-inverse solutions for joint velocity ${\dot{\bm \varTheta}_k}$ and acceleration ${\ddot{\bm \varTheta}_k}$ of each robot are
\begin{equation}\label{eqn:inverse_kinematics}
{\dot{\bm \varTheta}_k} = {\bm J_k^\dagger}{\dot{\bm x}_{k}}  \hspace{1.5mm} \text{and} \hspace{1mm} \ 
{{\ddot{\bm \varTheta}_k}={\bm J_k^\dagger}\left({\ddot{\bm x}_{k}}-{\dot{\bm J}_k} {\dot{\bm \varTheta}_k}\right)},
\end{equation}
respectively; where ${\bm J_k^\dagger}$ is the pseudo-inverse of $\bm J_k$.

In this paper, the null-space motion of the redundant manipulators is used to optimize the force manipulability
given the current WFME of robot $k$ and a desired nominal task ellipsoid (Section~\ref{sec:nominal_task_ellipsoid}).
By 
using the tensor representation and exploiting the fact 
that ellipsoids lie on the Riemannian manifold of symmetric positive definite (SPD) matrices~\cite{jaquier2018geometry},
the velocity-level inverse kinematics with task-oriented manipulability optimization can be derived as 
\begin{equation}\label{eqn:null_space_geometry_tracking}
\resizebox{0.89\hsize}{!}{$
{\dot{\bm \varTheta}_k}=\bm {J}_k^{\dagger} {\dot{\bm x}_{k}}+\left(\bm {I}-\bm{J}_k^{\dagger} \bm{J}_k\right)\left(\bm{\mathcal{J}}_{M_k}^{f\dagger}\right)^{T}  \bm K_M \operatorname{vec}\left(\dot{\bm M}_k^f\right).
$}
\end{equation}
The first term of the right hand side is same as in~\eqref{eqn:inverse_kinematics}, while the second term projects the difference between the nominal task ellipsoid and WFME to the null space motion of the manipulator. $\bm K_M$ is the scaling matrix, $\operatorname{vec} \left(\right)$ denotes the vectorization of symmetric matrices with Mandel notation\footnote{The Mandel representation $[\bm \varepsilon]$ (as a column-vector) of any second rank, symmetric tensor $\bm \varepsilon$ is defined as follows: $[\bm \varepsilon] = [\varepsilon_{11}, \varepsilon_{22}, \sqrt2\varepsilon_{12}]^{T}.$},  $\dot{\bm M}_k^f=\operatorname{Log}_{\bm{M}_k^{f}} {{\bm M}_{t}}$, the $\operatorname{Log}$ operator is a logarithm map~\cite{pennec2006riemannian} which can find the tangent vector between two points in the SPD manifold. ${\bm{M}_k^{f}}$ represents the WFME of robot $k$ and is defined as
\begin{equation}
    \bm M^f_k= \left( \bm J_k \bm W_k^T \bm W_k \bm J_k^T \right)^{-1}.
\end{equation}
The force manipulability Jacobian $\bm {\mathcal{J}}_{M_k}^f$ projects the scaled rate of change of $\bm M^f_k$ to the joint velocity ${\dot{\bm \varTheta}_k}$ and it is obtained as
\begin{equation}
\begin{split}
   \bm {\mathcal{J}}_{M_k}^f &=
-\dfrac{\partial {\bm M^{f-1}_k}}{\partial {\bm \varTheta_k}} 
\times_1 \bm M^f_k \times_2 \bm M^f_k, \\
\dfrac{\partial {\bm M^{f-1}_k}}{\partial {\bm \varTheta_k}} &=
 \dfrac{ {\partial \bm J_k}}{\partial {\bm \varTheta_k}}\times_2 \bm J_W + \dfrac{ {\partial \bm J_k^T}}{\partial {\bm \varTheta_k}}\times_1 \bm J_W,
\end{split}
\end{equation}
where $\bm J_W=\bm J_k\bm W_k^T \bm W_k$, $\times_n$ is the $n$-mode tensor product\footnote{$1$-mode and $2$-mode: $\bm A \times_1 \bm U^T=\bm U^T\bm A$, $\bm A \times_2 \bm V^T=\bm A \bm V$}~\cite{kolda2009tensor}.
The force capability optimization is achieved in the null space of manipulation task by using the proposed task-oriented manipulability optimization~\eqref{eqn:null_space_geometry_tracking}.

%% file: figures/force_polytope_illustration_index.tex
\begingroup%
  \makeatletter%
  \providecommand\color[2][]{%
    \errmessage{(Inkscape) Color is used for the text in Inkscape, but the package 'color.sty' is not loaded}%
    \renewcommand\color[2][]{}%
  }%
  \providecommand\transparent[1]{%
    \errmessage{(Inkscape) Transparency is used (non-zero) for the text in Inkscape, but the package 'transparent.sty' is not loaded}%
    \renewcommand\transparent[1]{}%
  }%
  \providecommand\rotatebox[2]{#2}%
  \newcommand*\fsize{\dimexpr\f@size pt\relax}%
  \newcommand*\lineheight[1]{\fontsize{\fsize}{#1\fsize}\selectfont}%
  \ifx\svgwidth\undefined%
    \setlength{\unitlength}{1445.66929134bp}%
    \ifx\svgscale\undefined%
      \relax%
    \else%
      \setlength{\unitlength}{\unitlength * \real{\svgscale}}%
    \fi%
  \else%
    \setlength{\unitlength}{\svgwidth}%
  \fi%
  \global\let\svgwidth\undefined%
  \global\let\svgscale\undefined%
  \makeatother%
  \begin{picture}(1,0.62745098)%
    \lineheight{1}%
    \setlength\tabcolsep{0pt}%
    \put(0,0){\includegraphics[width=\unitlength,page=1]{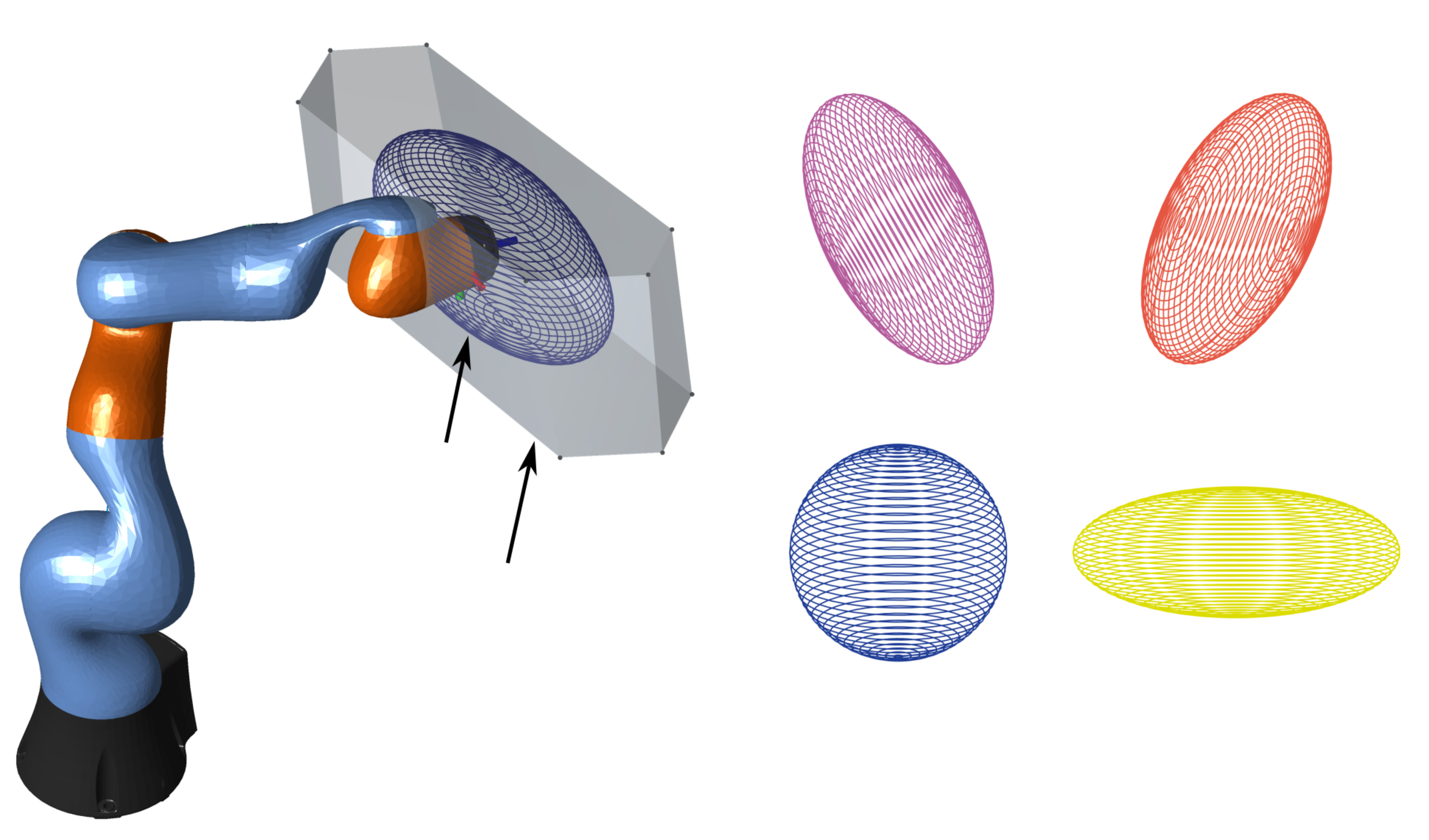}}%
    \put(0.49687081,0.08591902){\color[rgb]{0,0,0}\makebox(0,0)[lt]{\lineheight{1.0}\smash{\begin{tabular}[t]{c}\footnotesize{Isotropic ellipsoid}\\ \hspace{2mm}$\scriptstyle{c_f = 1}$\end{tabular}}}}%
    \put(0.785094031,0.08621902){\color[rgb]{0,0,0}\makebox(0,0)[lt]{\lineheight{1.0}\smash{\begin{tabular}[t]{c} $\scriptstyle{F_t =[3,0,0]}$\\ $\scriptstyle{c_f = 0.4}$ \end{tabular}}}}%
    \put(0.77094031,0.30746357){\color[rgb]{0,0,0}\makebox(0,0)[lt]{\lineheight{1.25}\smash{\begin{tabular}[t]{l}$\scriptstyle{F_t =[3,4,5]}$\end{tabular}}}}%
    \put(0.53871455,0.30746357){\color[rgb]{0,0,0}\makebox(0,0)[lt]{\lineheight{1.25}\smash{\begin{tabular}[t]{l}$\scriptstyle{F_t =[-3,4,5]}$\end{tabular}}}}%
    \put(0.22086344,0.24540715){\color[rgb]{0,0,0}\makebox(0,0)[lt]{\lineheight{1.25}\smash{\begin{tabular}[t]{l}\small{WFME}\end{tabular}}}}%
    \put(0.28394902,0.16845955){\color[rgb]{0,0,0}\makebox(0,0)[lt]{\lineheight{1.0}\smash{\begin{tabular}[t]{c}\small{Force}\\  \small{polytope}\end{tabular}}}}%
    \put(0.6905463,0.48940178){\color[rgb]{0,0,0}\makebox(0,0)[lt]{\lineheight{1.25}\smash{\begin{tabular}[t]{l}$\scriptstyle{c_f = 0.5}$\end{tabular}}}}%
  \end{picture}%
\endgroup%

%% file: sections/A3C.tex
According to the force polytope shown in Fig.~\ref{fig:singlearm_illustration}, the maximum operational force along any specific direction can be calculated. 
Subsequently, the decentralized ability-aware adaptive controller ($\textit{DA}^3\textit{C}$)  
computes the control inputs in accordance with the force capability of each robot.

\subsection{Force Capability}\label{sec:maximum_force}
The maximum operational force of each manipulator along the specific task direction $\bm F_t$ is calculated by 
\begin{IEEEeqnarray}{CC}
    \IEEEyesnumber\label{eq:maximum_force} 
    \IEEEyessubnumber \label{eq:max_force}
    \max_{{\bm F_{k}}} \ & \left\|{\bm F_{k}}\right\| _2\\
    ~\text{s.t.}~  
    \IEEEyessubnumber \label{eqn:dynamics_constraint} &
   { |{\bm H_k}{ \ddot{\bm \varTheta}_k}+{\bm C_k}+{\bm G_k} +  {\bm J_k}^T{\bm F}_{k} | \leq {\bm \tau_{k,max}}}~\hfill\\
    \IEEEyessubnumber \label{eqn:direction_constraint} & 
    \bm F_{t}^{\times} {\bm F}_{k} = {\bm 0}
    , \hfill
\end{IEEEeqnarray}
where ${\bm H_k}$ is the inertia matrix of {robot $k$}, ${\bm C_k}$ is the Coriolis and centrifugal force of {robot $k$}, ${\bm G_k}$ is the gravity of {robot $k$}. The force polytope is defined by \eqref{eqn:dynamics_constraint} and can be obtained from the dynamic equation of the manipulator, while \eqref{eqn:direction_constraint} is used to define the specific task direction. 

%
%

\subsection{Ability-Aware Adaptive Controller}

In order to track the desired trajectory of the object during decentralized multi-robot collaborative manipulation, we propose an ability-aware adaptive controller in which the force capability of each robot is considered. 


According to \eqref{eqn:target_dynamic_equation}, the ideal reference dynamics model of the object can be written as
\begin{equation}\label{eqn:reference_model}
{\ddot {\bm x}_o^*} = \bm A^* {\dot {\bm x}_o^*} + \bm B^* \left(\bm F_t^* - \bm N_{cg} \right),
\end{equation}
where $\bm A^*$ is a Hurwitz stable matrix written as $\bm A^*=\left[\begin{array}{cc} -\frac{\mu_l}{m_o^*} \bm E_3 & {\bm 0} \\ {\bm 0} & -\mu_r \bm I_o^{*-1}  \end{array} \right]$, $\bm B^*=\left[\begin{array}{cc} \frac{1}{m_o^*} \bm E_3 & {\bm 0} \\ {\bm 0} & \bm I_o^{*-1}  \end{array} \right]$, and $\bm N_{cg}=\left[\begin{array}{c}  -m_o^* \bm g  \\ \bm \omega_o \times({\bm I_o^*}{\bm \omega_o}) \end{array} \right]$ is a stacked vector of gravity term and nonlinear term in~\eqref{eqn:target_dynamic_equation}, and $m_o^*$ and $\bm I_o^*$ are the nominal mass and inertia matrix of the object. Given the bounded reference (desired) trajectory ${\dot {\bm x}_o^*}$ and ${\ddot {\bm x}_o^*}$ of the object, the bounded reference control input $\bm F_t^*$, which represents the external force exerted on the object, can be computed from~\eqref{eqn:reference_model}.

To design the adaptive controller, the actual object's dynamics model 
can be rewritten in the following linear form with respect to the system state ${\dot {\bm x}_o}$ as
\begin{equation}\label{eqn:actual_model}
 \begin{split}
	{\ddot {\bm x}_o} &= \bm A {\dot {\bm x}_o} +  \sum_{k=1}^K \bm B_k \left(\bm F_k - \bm U_{k} \right), \\
 \end{split}
\end{equation}
where $\bm A$ and $\bm B_k$ are unknown constant matrices, $\bm A=-\left[\begin{array}{cc} \frac{\mu_l}{m_o} \bm E_3 & {\bm 0} \\ {\bm 0} & \mu_r \bm I_o^{-1}  \end{array} \right]$, $\bm B_k=\left[\begin{array}{cc} \frac{1}{m_o} \bm E_3 & {\bm 0} \\ \bm I_o^{-1} \bm r_k^{\times} & \bm I_o^{-1}  \end{array} \right]$, $\bm F_k$ is the input of robot $k$, $K$ is the total number of robots, $\bm U_{k} = \bm W_{\phi k}^{*T} \bm \varPhi_k$ is an unknown nonlinear term caused by modelling uncertainties which can be approximated by a Radial Basis Functions (RBFs) neural network, $\bm W_{\phi k}^*$ is the weight matrix for the RBFs, $\bm \varPhi_k$ is the vector of RBFs and output bias.


For each robot, the adaptive control input is designed as
\begin{equation}
	\bm F_k = \bm K_{xk}^T {\dot{\bm x}_o} + \bm K_{rk}^T \bm F_t^* + \bm K_{nk}^{T} \bm N_{cg} + {\bm W}_{\phi k}^T \bm \varPhi_k,
\end{equation}
where $\bm K_{xk}$, $\bm K_{rk}$, $\bm K_{nk}$ and ${\bm W}_{\phi k}$ are control gain matrices. 
Considering the force capability of each robot, the control input constraints of $\textit{DA}^3\textit{C}$ are guaranteed by positive $\mu$-modification~\cite{lavretsky2004positive}. 
Therefore, 
the modified ideal reference dynamics model (see~\eqref{eqn:reference_model}) and the actual system dynamics (see~\eqref{eqn:actual_model}) with input constraints are rewritten as
\begin{equation}\label{eqn:modified reference model}
{\ddot {\bm x}_o^*} = \bm A^* {\dot {\bm x}_o^*} + \bm B^* \left(\bm F_t^* +\sum_{k=1}^{K} \bm K_{fk}^T \Delta\bm F_{k} - \bm N_{cg} \right), 
\end{equation}
\begin{equation}\label{eqn:constrained system dynamics}
{\ddot {\bm x}_o} = \bm A {\dot {\bm x}_o} + \sum_{k=1}^K \bm B_k \left( \bm F_{k} +\Delta\bm F_{k} - \bm U_{k} \right), 
\end{equation}
where $\bm K_{fk}$ is the control gain matrix, $\Delta\bm F_{k}$ is the control deficiency, which is described in Appendix A, and is used to generate the adaptive augmentation for the reference model. 



\begin{figure}[htb]
    \centering
     \begin{subfigure}[b]{0.240\textwidth}
         \centering
         \includegraphics[width=\textwidth]{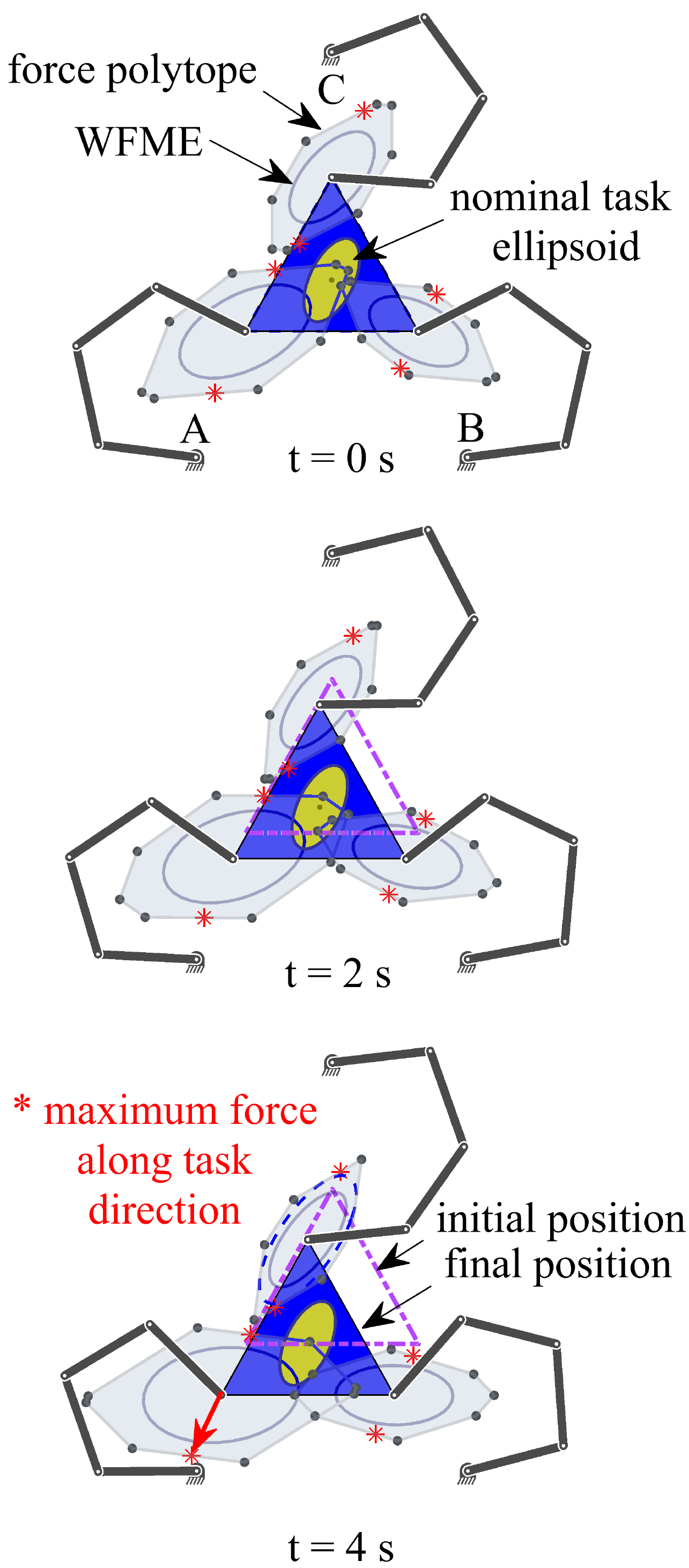}
         \caption{Without optimization}
         \label{fig:position}
     \end{subfigure}
     \hfill
    \begin{subfigure}[b]{0.205\textwidth}
         \centering
         \includegraphics[width=\textwidth]{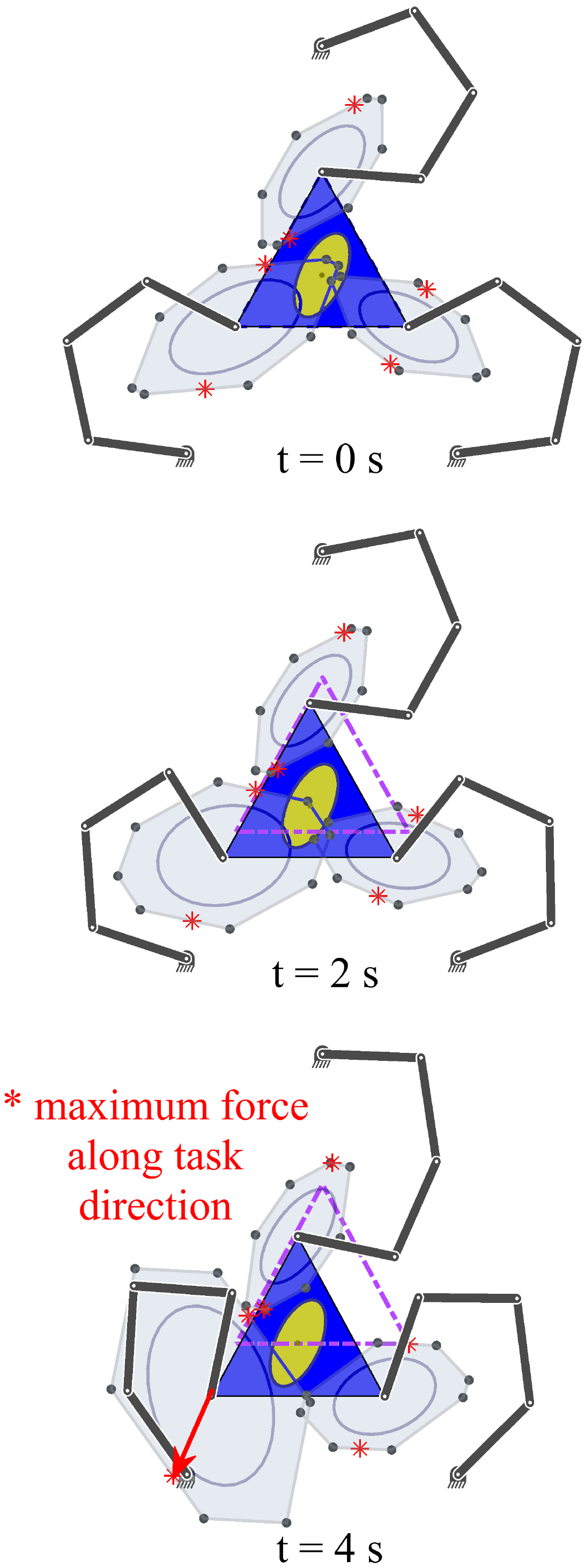}
         \caption{With optimization}
         \label{fig:position}
     \end{subfigure}
	\caption{Simulation results with and without manipulability optimization. }\label{fig:sim_mo}
    \vspace{-15pt}
\end{figure}

Furthermore, by defining the tracking error $\bm e=\dot{\bm x}_o-\dot{\bm x}_o^*$, the tracking error dynamics can be obtained as 
\begin{equation}\label{eqn:error_dynamics}
\begin{split}
&\dot{\bm e} = \ddot{\bm x}_o-\ddot{\bm x}_o^* = \bm A^* {\bm e} -\sum_{k=1}^{K} \bm B^* \tilde{\bm K}_{fk}^{T} \Delta\bm F_{k} + \\
            &\sum_{k=1}^{K} \bm B_k \left( \tilde{\bm K}_{xk}^{T} \dot{\bm x}_o +\tilde{\bm K}_{rk}^{T} \bm F_t^* +\tilde{\bm K}_{nk}^{T} \bm N_{cg} + \tilde{\bm W}_{\phi k}^{T} \bm \varPhi_k \right), 
\end{split}
\end{equation}
where $\tilde{\bm K}=\bm K -\bm K^*$ is the error matrix of control gain and $\bm K^*$ is ideal control gain (see Appendix B).

Last, the following adaptive laws are adopted in order to guarantee asymptotic tracking of the reference trajectory. 
\begin{equation}\label{eqn:adaptive_law}
\begin{split}
	\dot{\tilde{\bm K}}_{xk} &= -\bm \varGamma_{xk}\dot{\bm x}_o \bm e^T \bm P \bm B_k,
	\dot{\tilde{\bm K}}_{rk} = -\bm \varGamma_{rk}\bm F_t^* \bm e^T \bm P \bm B_k, \\
	\dot{\tilde{\bm K}}_{nk} &= -\bm \varGamma_{nk} \bm N_{cg} \bm e^T \bm P \bm B_k,
	\dot{\tilde{\bm W}}_{\phi k} = -\bm \varGamma_{\phi k} \bm \varPhi \bm e^T \bm P \bm B_k, \\
	\dot{\tilde{\bm K}}_{fk} &= \bm \varGamma_{fk} \Delta \bm F_k \bm e^T \bm P \bm B^*,
\end{split}
\end{equation}
where $\bm P$ is unique SPD solution of the Lyapunov equation, $\bm P \bm A^*+ \bm A^{*T} \bm P = -\bm Q < \bm 0$, where $\bm Q$ is any SPD matrix; $\bm \varGamma_{xk}$, $\bm \varGamma_{rk}$, $\bm \varGamma_{nk}$, $\bm \varGamma_{\phi k}$, and $\bm \varGamma_{fk}$ are SPD gain matrices. The proof for asymptotic stability and asymptotic tracking using the above adaptive laws can be found in Appendix B.
\subsection{Computed-Torque Control for Each Robot}
According to the constrained operational force and the adaptive reference trajectory of the object, the controller of each robot can be designed as the following well known computed-torque feedforward control, 
\begin{equation}\label{eqn:manipulator dynamics}
\begin{split}
{\bm \tau_k} =& {\bm H_k} \left({ \ddot{\bm \varTheta}_k} + \bm K_p \tilde{\bm \varTheta}_k + \bm K_v {\dot{\tilde{\bm \varTheta}}_k} \right) \\
              & +{\bm C_k} +{\bm G_k} + {\bm J_k}^T\left( \bm F_{k} +\Delta\bm F_{k} \right),
\end{split}
\end{equation}
where $\tilde{\bm \varTheta}_k$, $\dot{\tilde{\bm \varTheta}}_k$ are the error vectors of joint angle and joint angular velocity, and $\bm K_p$, $\bm K_v$ are the corresponding gains.
 

\subsection{Level of Communication}



In the proposed ability-aware adaptive controller, the levels of communication between the robots in the team may vary depending on the different states of the multi-robot system. 
As long as the desired operational force of each robot is within its force polytope, each robot can be controlled in a fully decentralized manner.
Each robot should know the object's velocity and the grasping location on the object. However, in cases where a robot would need to exceed its capabilities to track the reference input---desired operational force lies outside of the force polytope---then, the control deficiency  $\Delta\bm F_{k}$ 
of each robot should be broadcast to all the robots of the team. The control deficiency of all the robots will be used in~\eqref{eqn:modified reference model} to modify the unfeasible reference control input for each robot, which results in a coordinated adaptation of the robot team.

%% file: sections/results.tex
First, we perform three ablation studies to demonstrate the benefits of each one of the proposed components and their combination.
Second, we demonstrate the adaptation capabilities of the $\textit{DA}^3\textit{C}$ framework in a decentralized manipulation setup. Third, we validate $\textit{DA}^3\textit{C}$ on a physical simulation, where three heterogeneous robots manipulate an object. The control frequency of all the simulations is set to 500 Hz.

\subsection{Ablation Studies}

\subsubsection{Force Capabilities with and without Manipulability Optimization}\label{sec:sim_mo}



In this ablation study, three 4-DOF planar robots manipulate an object from initial position [0, 0] $\si{m}$ to final position [-0.1, -0.2] $\si{m}$.  The maximum joint torques of robot A, B and C are set to 0.8, 0.6 and 0.6 $\si{N.m}$, respectively.
The simulation results without and with task-oriented manipulability optimization are shown in~\cref{fig:sim_mo}. 
Using the proposed method the WFME of each robot is optimized through the null-space motion to track the nominal task ellipsoid (yellow ellipsoid), and consequently the force capability (red arrow) along the specific task direction is increased.




\begin{figure*}[!htb]
    \vspace{1.5mm}
	\begin{minipage}{0.32\textwidth}
		\centering
    	\includegraphics[width=0.98\textwidth]{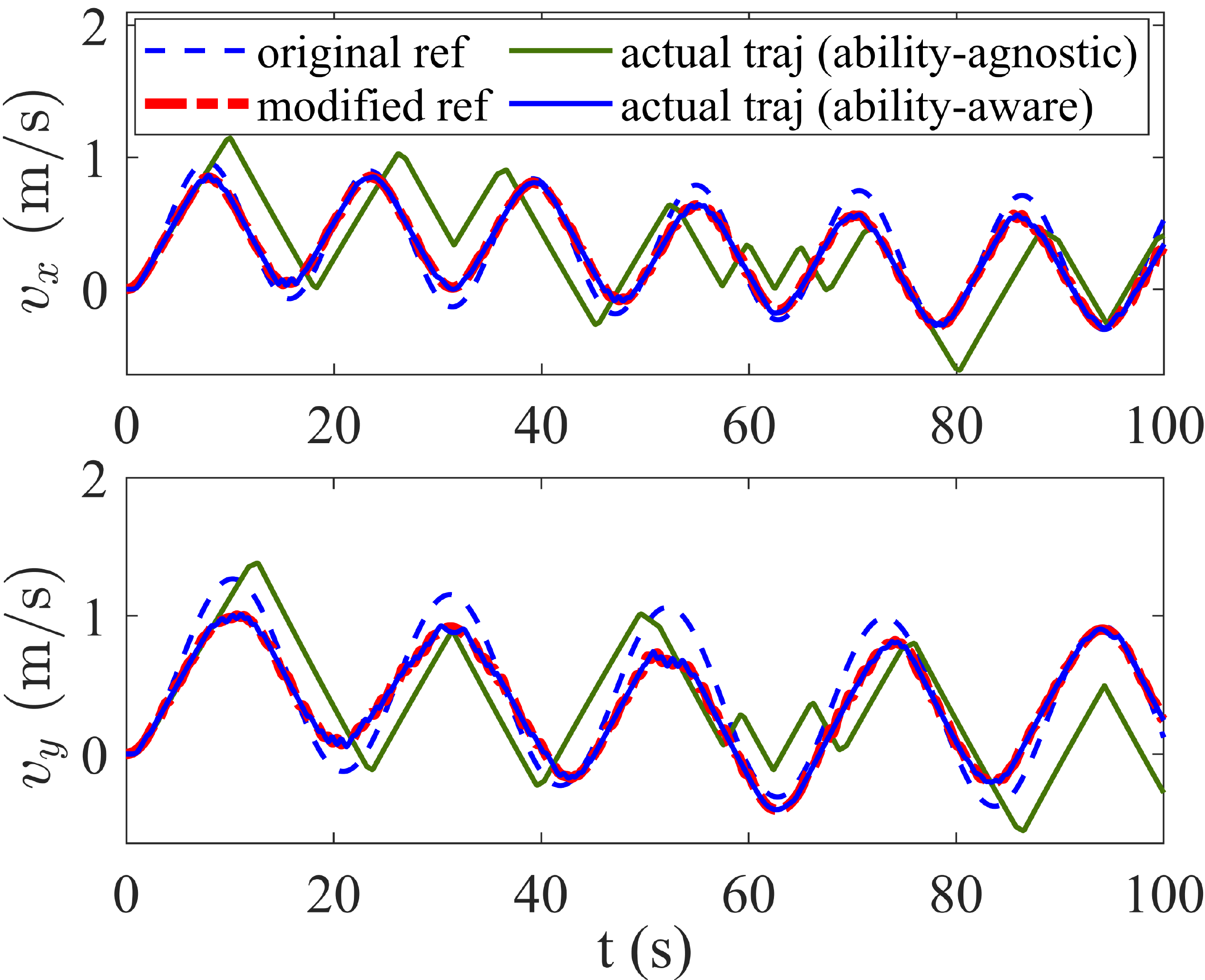}
    	\vspace{-5pt}
    	\caption{State of the object with ability-agnostic and ability-aware controller.}\label{fig:state_traj_wwo}
    	\vspace{-12pt}
	\end{minipage}\hfill
    \begin{minipage}{0.32\textwidth}
		\centering
    	\includegraphics[width=0.98\textwidth]{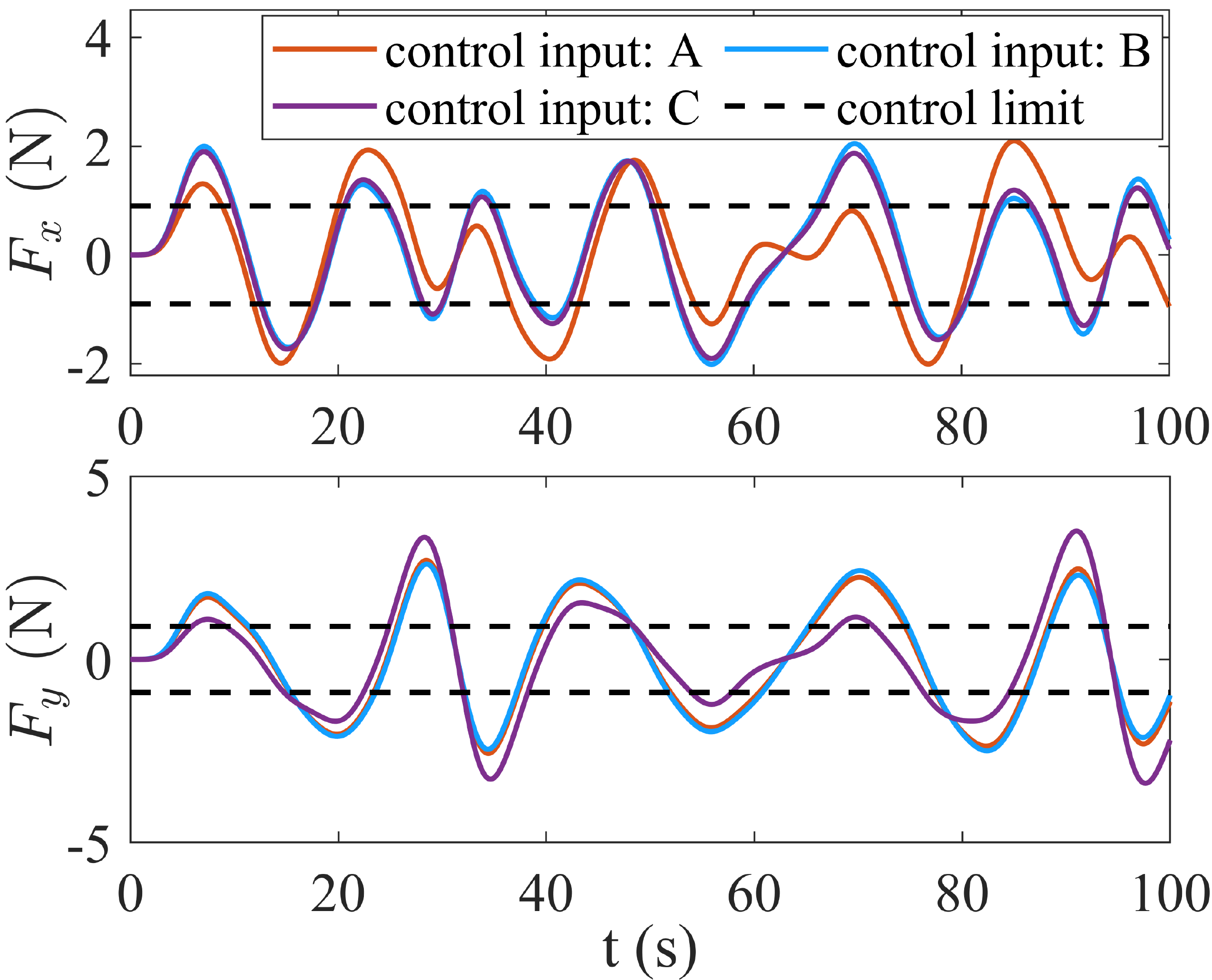}
    	\vspace{-5pt}
    	\caption{Control input of each robot with ability-agnostic controller.}\label{fig:control_traj_wo}
    	\vspace{-12pt}
	\end{minipage}\hfill
	\begin{minipage}{0.32\textwidth}
		\centering
    	\includegraphics[width=
    	0.98\textwidth]{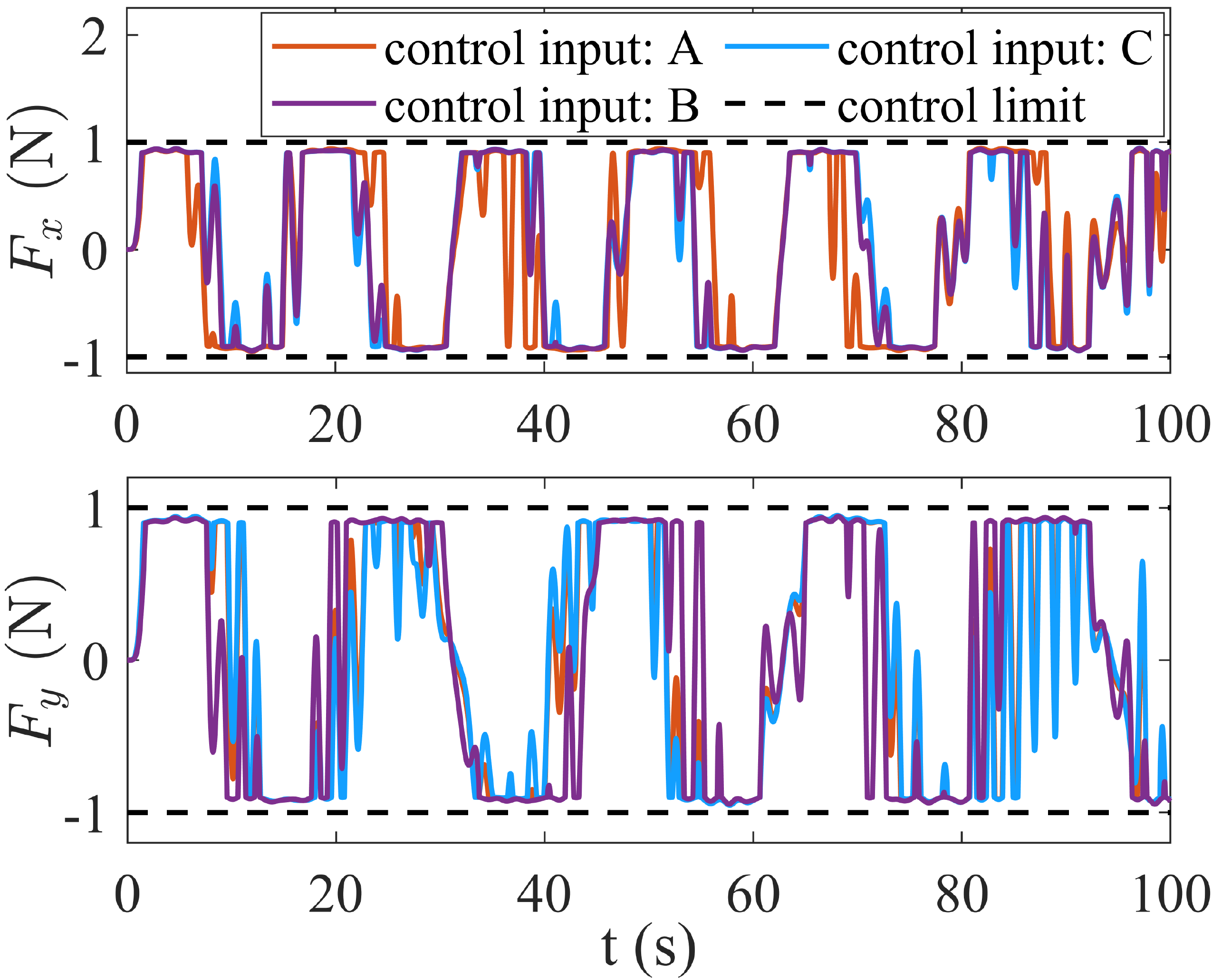}
    	\vspace{-5pt}
    	\caption{Control input of each robot with ability-aware controller.}\label{fig:control_traj_w}
    	\vspace{-12pt}
	\end{minipage}
\end{figure*}

\begin{figure*}[!htb]
	\centering
	\includegraphics[width=0.98\textwidth]{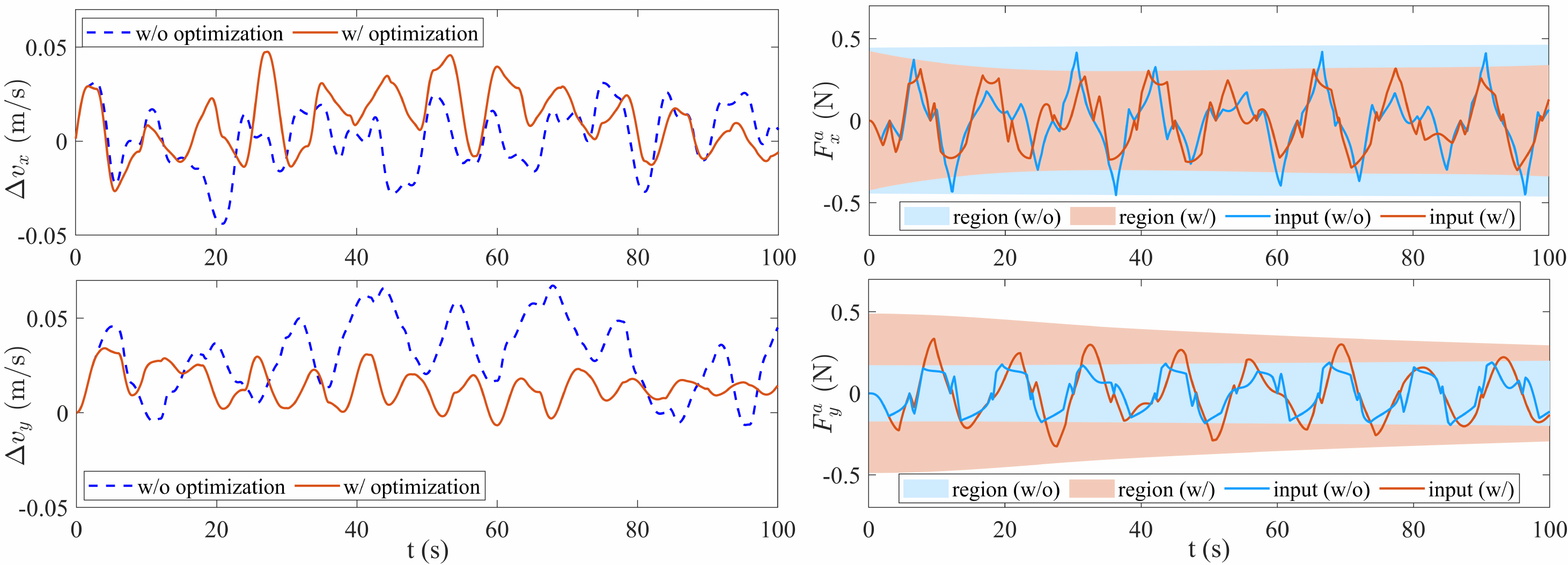}
	\vspace{-5pt}
	\caption{Tracking errors of the object (1st column) and control inputs of robot A (2nd column) with and without manipulability optimization. The cyan regions represent the force capability envelope of the robot without optimization while the orange regions represents the capability envelope with optimization. }\label{fig:state_control_mo_new}
	\vspace{-10pt}
\end{figure*}

\begin{figure*}[htb]
	\centering
	\includegraphics[width=0.95\textwidth]{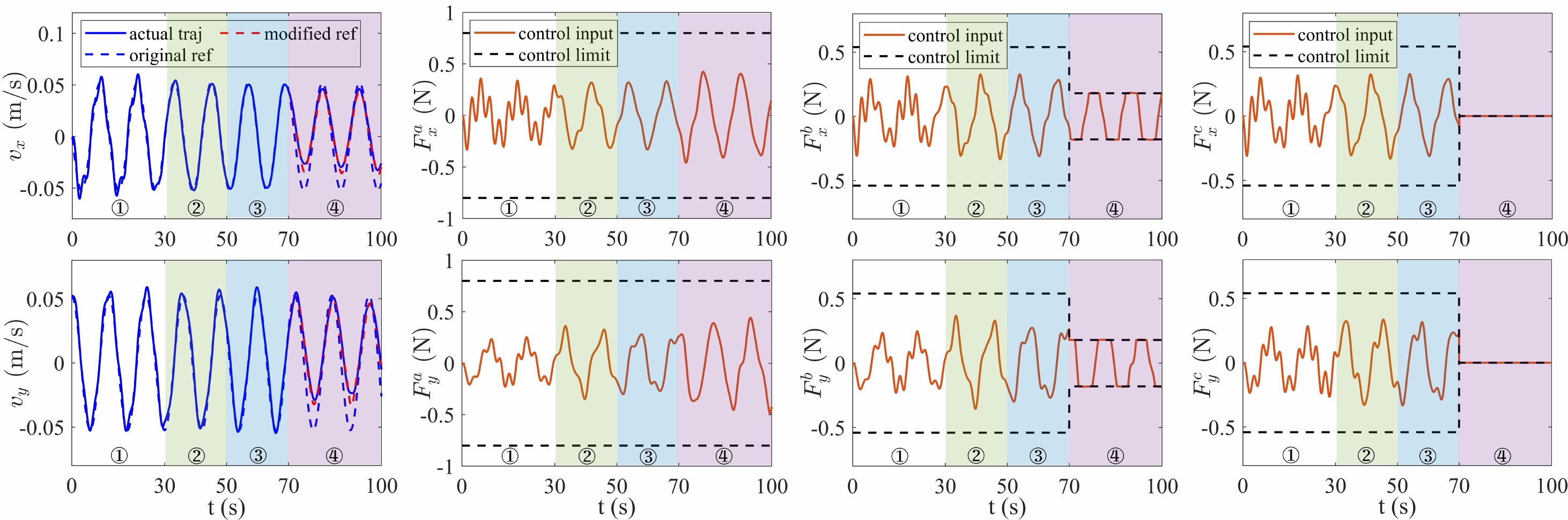}
	\vspace{-5pt}
	\caption{State of the object (1st column) and control inputs of robot A (2nd column) and robot B (3rd column) and robot C (4th column). In region \raisebox{.5pt}{\textcircled{\raisebox{-.9pt} {$2$}}}, the mass of the object is increased from 20 kg to 30 kg. In region \raisebox{.5pt}{\textcircled{\raisebox{-.9pt} {$3$}}}, the friction coefficient is decreased from 0.2 to 0.1, while in region \raisebox{.5pt}{\textcircled{\raisebox{-.9pt} {$4$}}}, the force capability of robot B is reduced to 1/3 while robot C is shut off.}\label{fig:da3c_test_new}
	\vspace{-12pt}
\end{figure*}

\subsubsection{Ability-Aware vs Ability-Agnostic Adaptive Controller}\label{sec:sim_aa}
The second ablation study compares the tracking performance of the proposed ability-aware adaptive controller ($\textit{DA}^3\textit{C}$) with  an ability-agnostic adaptive controller~\cite{culbertson2018decentralized}. 
The task considers three robots manipulating an object, with $\SI{20}{kg}$ mass and $\SI{20}{kg.m^2}$ inertia (along z-axis), on a plane with a sliding friction coefficient of 0.2. The adaptive controllers are initiated with 80$\%$ of these values. The reference control input is set to $\bm F_t^*=4*[\sin(0.4t), \ \sin(0.3t)]^T \ \si{N}$. The maximum force $\bm F_{max}$ of each robot is set to [1.0, 1.0] $\si{N}$, and the constant vector $\bm \delta$ (see Appendix A) is set to 10$\%$ of $\bm F_{max}$. The object's trajectories with ability-agnostic control (green line) and ability-aware control (blue line) are shown in~\cref{fig:state_traj_wwo}. 
For the ability-agnostic adaptive controller, the control input (see~\cref{fig:control_traj_wo}) exceed the force constraints,
hence, we limit the control input to the maximum force.
This results in a significant deviation of the object's trajectory from the desired one (see~\cref{fig:state_traj_wwo}).
On the other hand, 
given the limited capability of each robot in the ability-aware controller (see~\cref{fig:control_traj_w}) the new reference trajectory (red dashed line) deviates from the original one (blue dashed line) according to the control deficiency of all the robots as shown in~\cref{fig:state_traj_wwo}. In this way, 
the multi-robot system can track the modified trajectory accurately with average tracking error 0.009 $\si{m/s}$ and 0.013 $\si{m/s}$ along x-axis and y-axis, respectively.

\begin{figure*}[htb]	
	\centering
	\includegraphics[width=0.99\textwidth]{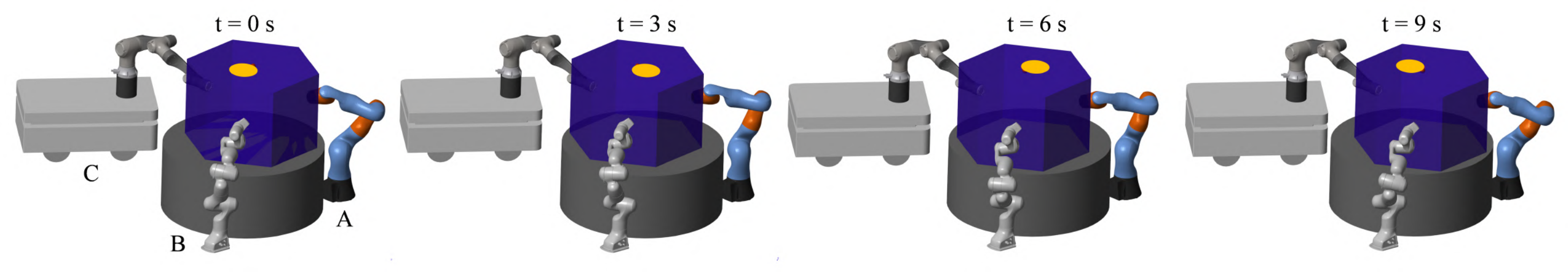}
	\vspace{-5pt}
	\caption{A heterogeneous multi-robot team manipulates an object. The object's circular motion is illustrated with respect to the yellow circle, which is fixed in the world frame.}\label{fig:sim_A3C}
	\vspace{-10pt}
\end{figure*}

\begin{figure*}[htb]	
	\centering
    \includegraphics[width=0.98\textwidth]{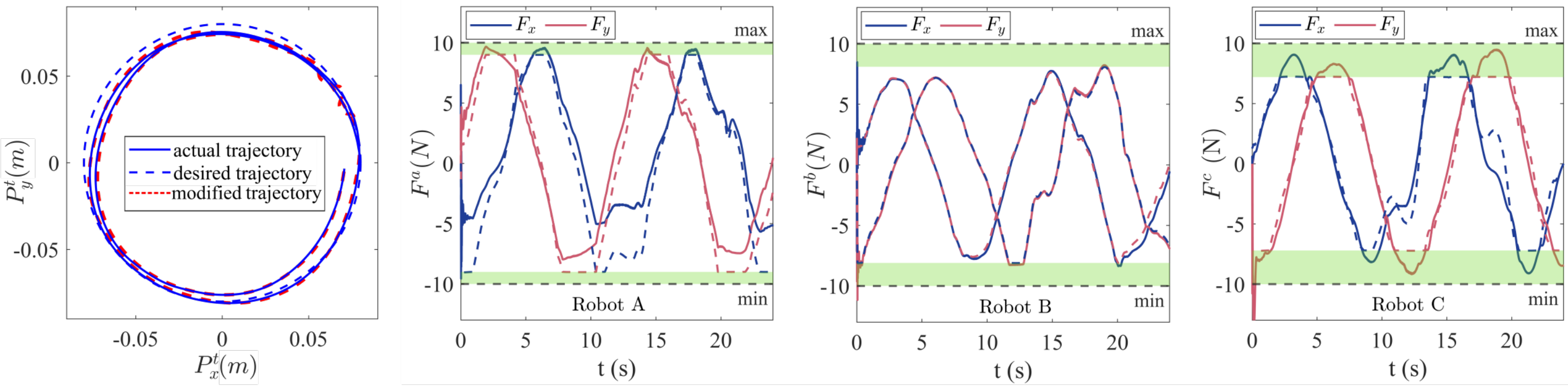}	
	\vspace{-5pt}
	\caption{Desired, modified reference and actual trajectory of the object (1st column), and desired (dashed line) and actual (solid line) end-effector forces of robot A (2nd column), robot B (3rd column) and robot C (4th column). By setting robot-specific safety zones (green region), the force of each robot is guaranteed not to violate its actual limits after a few iterations.}\label{fig:physical_state_control}
	\vspace{-10pt}
\end{figure*}

\subsubsection{$\textit{DA}^3\textit{C}$ with and without Manipulability Optimization}\label{sec:DAwwomanip}
Here, we compare the proposed  ability-aware adaptive control with and without manipulability optimization. The three manipulators of~\cref{sec:sim_mo}) are used to manipulate the same object as in~\cref{sec:sim_aa}). 
The reference trajectory is circular and is described with $v_x = -0.1*\frac{\pi}{6}*\sin(\frac{\pi}{6}t)$ $\si{m/s}$, $v_y = 0.1*\frac{\pi}{6}*\cos(\frac{\pi}{6}t)$ $\si{m/s}$ and $w_z = 0$. 
The nominal task ellipsoid is a sphere to equally allot the force capabilities along different directions.
The velocity tracking errors of the object are shown in~\cref{fig:state_control_mo_new} (left). The average tracking errors along two axes with manipulability optimization are (0.0138, 0.0143) $\si{m/s}$, which outperforms (0.0138, 0.0301) $\si{m/s}$ corresponding to the one without. In the right part of~\cref{fig:state_control_mo_new}, we can observe all the control inputs along with the force limits (regions). The $\textit{DA}^3\textit{C}$ with manipulability optimization regulates the robot's configuration to achieve similar force capabilities
along both directions, such that tracking performance is balanced in both directions.


\subsection{$\textit{DA}^3\textit{C}$ on-the-fly Adaptation}

Here, we show that the proposed method is capable of handling changes with respect to the mass of the object, the friction coefficient and is also tolerant to faults of team members. We consider the same circular task, where the mass of the object is increased from 20 $\si{kg}$ (white region \raisebox{.5pt}{\textcircled{\raisebox{-.9pt} {$1$}}}) to 30 $\si{kg}$ (green region \raisebox{.5pt}{\textcircled{\raisebox{-.9pt} {$2$}}}) at 30 $\si{s}$, the friction coefficient is decreased from 0.2 (green region \raisebox{.5pt}{\textcircled{\raisebox{-.9pt} {$2$}}}) to 0.1 (blue region \raisebox{.5pt}{\textcircled{\raisebox{-.9pt} {$3$}}}) at 50 $\si{s}$, and the max force of robot B is reduced to 1/3 while robot C is shut off (purple region \raisebox{.5pt}{\textcircled{\raisebox{-.9pt} {$4$}}}) at 70 $\si{s}$. 
The results are shown in ~\cref{fig:da3c_test_new}, the control inputs of the robots increase when the object's mass is increased and decrease when the friction coefficient is decreased. Further, it is worth noting that in the purple region \raisebox{.5pt}{\textcircled{\raisebox{-.9pt} {$4$}}}, the control input of robot A is increased and the reference trajectory is adapted, due to the loss of robot C and the reduced capability of robot B.

\subsection{$\textit{DA}^3\textit{C}$ for Multi-Robot Collaboration}
To verify the proposed $\textit{DA}^3\textit{C}$ framework, three heterogeneous robots, the torque-controlled Kuka-iiwa (A), the Franka-panda (B), and one position-based admittance controlled mobile manipulator (C), are used to  manipulate an object on a surface (see in~\cref{fig:sim_A3C}), where both the object's mass, and the friction between the object and the surface are unknown.
The task is a circular motion
and the maximum force for all the robots is limited to 10 $\si{N}$, while the constant $\bm \delta$ (see Appendix A) are set to 1, 2, and 3 $\si{N}$, respectively. For the torque-controlled robots, the feedback gains $\bm K_p$, $\bm K_v$ in~\eqref{eqn:manipulator dynamics} are set to $\operatorname{diag}$([16, 16, 16, 12, 12, 12, 12]) and $\operatorname{diag}$([0.2, 0.2, 0.2, 0.1, 0.1, 0.1, 0.1]), respectively. The interaction between the object and the robots is modelled with a spring-damper (stiffness 5000 $\si{N/m}$ and damping 500 $\si{Ns/m}$). 
The physical simulation and visualization are developed in MATLAB Simscape and Simulink. 

The object's desired and actual trajectories, and the force profiles of each robot are shown in ~\cref{fig:physical_state_control}. 
The reference trajectory is modified according to the control deficiency, due to the limits on the robots' force capabilities.
Robot C does not always track the desired force accurately, due to the admittance controller. Thus, its safety zone is set larger (green area) to always experience forces within its limits. Also, such force deviations introduce further unmodelled interaction forces, due to the formed closed-chain. Yet, each robot adapts its control gains on-the-fly to cope with these deviations too.

%% file: sections/conclusion.tex

In this paper we propose a decentralized ability-aware adaptive control ($DA^3C$) framework for multi-robot collaborative manipulation, which can handle uncertain system parameters, input constraints and band-limited communication. The key idea is that the force capability of each robot is maximized by exploiting its null-space motion, while the designed adaptive controller enables decentralized coordination according to the capability of each robot.
The proposed method achieves accurate trajectory tracking irrespective of the low-level controllers, and can be used for heterogeneous fixed-base and mobile-base multi-robot systems. An open challenge is the inclusion of joint position limits into the ability-aware adaptive controller. 
In our future work, we plan to use $DA^3C$ for human-robot co-manipulation experiments where the access to human's capability is not straightforward.


%% file: sections/appendix.tex
\section{Control input with $\mu$-modification}\label{sec:mu_modification}
The control input with $\mu$-modification is written as
\begin{equation}
\begin{split}
	\bar{\bm F}_{k} &= \frac{1}{1+\mu} \left( \bm F_k+\mu \bm F_{kmax}^{\delta} \text{sat}\left( \frac{\bm F_k}{\bm F_{kmax}^{\delta}}\right) \right)\\
	&= { \begin{cases}&\bm F_k, \quad \left| \bm F_k \right| \leqslant \bm F_{kmax}^{\delta}  \\&\frac{1}{1+\mu} \left( \bm F_k +\mu \bm F_{kmax}^{\delta} \right), \quad \bm F_k  > \bm F_{kmax}^{\delta} \\&\frac{1}{1+\mu} \left( \bm F_k -\mu \bm F_{kmax}^{\delta} \right), \quad  \bm F_k  < -\bm F_{kmax}^{\delta} \end{cases} } ,
\end{split}
\end{equation}
where $\mu$ is positive design constant, $\bm F_{kmax}^{\delta}=\bm F_{kmax}-\bm \delta$, $\bm \delta$ is a constant vector, ${\bm 0}<\bm \delta< \bm F_{kmax}$, and $\bm F_{kmax}$ is the maximum force of robot $k$ which is obtained from~\eqref{eq:maximum_force}.
The corresponding control deficiency can be calculated as
\begin{equation}
	\Delta\bm F_{k}=\bm F_{kmax} \text{sat}\left(\frac{\bar{\bm F}_{k}}{\bm F_{kmax}}\right)-\bm F_k.
\end{equation}

\section{Lyapunov stability analysis}\label{sec:lyapunov_stability}
In order to match \eqref{eqn:modified reference model} and \eqref{eqn:constrained system dynamics}, we can choose the ideal gain matrices ${\bm K}_{xk}^*$,  ${\bm K}_{rk}^*$, ${\bm K}_{fk}^*$, and ${\bm K}_{nk}^*$ according to the following forms:
\begin{equation}
\begin{split}
\bm A + \sum_{k=1}^{K}\bm B_k \bm K_{xk}^{*T} = \bm A^*,& \quad \sum_{k=1}^{K}\bm B_k \bm K_{rk}^{*T} = \bm B^*,\\
-\sum_{k=1}^{K}\bm B_k \bm K_{nk}^{*T} = \bm B^*,& \quad \bm B^* \bm K_{fk}^{*T} = \bm B_k.
\end{split}
\end{equation}

According to the error dynamics in~\eqref{eqn:error_dynamics}, we consider the following Lyapunov function candidate:
\begin{equation}
\begin{split}	&V\left( \bm e, \tilde{\bm K} \right)= \bm e^T \bm P \bm e + \sum_{k=1}^{K} \text{tr}\left( \tilde{\bm K}_{xk}^T \bm \varGamma_{xk}^{-1}\tilde{\bm K}_{xk} \right)\\&+ \sum_{k=1}^{K} \text{tr}\left( \tilde{\bm K}_{rk}^T \bm \varGamma_{rk}^{-1}\tilde{\bm K}_{rk} \right) +\sum_{k=1}^{K} \text{tr}\left( \tilde{\bm K}_{nk}^T \bm \varGamma_{nk}^{-1}\tilde{\bm K}_{nk} \right) \\&+ \sum_{k=1}^{K} \text{tr}\left(\tilde{\bm W}_{\phi k}^T \bm \varGamma_{\phi k}^{-1}\tilde{\bm W}_{\phi k} \right) + \sum_{k=1}^{K} \text{tr}\left(\tilde{\bm K}_{fk}^T \bm \varGamma_{fk}^{-1}\tilde{\bm K}_{fk} \right).
\end{split}
\end{equation}
By using the adaptive law~\eqref{eqn:adaptive_law}, the derivative of Lyapunov function\footnote{For two real column matrices $\bm a \in \mathbb{R}^n$ and $\bm b \in \mathbb{R}^n$: $\text{tr} \left(  \bm b \bm a^T\right)=\bm a^T \bm b$.} decreases along the tracking error dynamics as
\begin{equation}
	\dot V \left( \bm e, \tilde{\bm K}_{xk}, \tilde{\bm K}_{rk}, \tilde{\bm K}_{nk}, \tilde{\bm K}_{fk}, \tilde{\bm W}_{\phi k} \right)= -\bm e^T \bm Q \bm e \leqslant 0.
\end{equation} 
Therefore, given a bounded reference input, we can conclude that the system can achieve asymptotic tracking by using Barbalat's Lemma. 